\documentclass{ws-us}
\usepackage[sort,compress,super]{cite}
\usepackage{amsmath}
\usepackage{float}

\begin{document}
\setlength{\pdfpagewidth}{8.5in}
\setlength{\pdfpageheight}{11in}
\catchline{0}{0}{2022}{}{}

\markboth{Xiangri LU,Zhanqing WANG, Hongbin MA}{Analysis of OODA Loop based on Adversarial for Complex Game Environments}

\title{Analysis of OODA Loop based on Adversarial for Complex Game Environments\footnote{The paper was supported by the National Natural Science Foundation of China(No.~62076028).}}

\author{Xiangri LU$^{a,b}$, Hongbin MA$^{a,b,}$\footnote{Corresponding author.}\ , Zhanqing WANG$^{a,b}$}

\address{$^a$State Key Laboratory of Intelligent Decision and Control for Complex Systems,\\ Beijing Institute of Technology, Beijing, China}

\address{$^b$Automated Institute, Beijing Institute of Technology, Beijing, China\footnote{The E-mail of corresponding author : mathmhb@139.com}}

\maketitle

\begin{abstract}
To address the problem of imperfect confrontation strategy caused by the lack of information of game environment in the simulation of non-complete information dynamic countermeasure modeling for intelligent game, the hierarchical analysis game strategy of confrontation model based on OODA ring (Observation, Orientation, Decision, Action) theory is proposed. At the same time, taking into account the trend of unmanned future warfare, NetLogo software simulation is used to construct a dynamic derivation of the confrontation between two tanks. In the validation process, the OODA loop theory is used to describe the operation process of the complex system between red and blue sides, and the four-step cycle of observation, judgment, decision and execution is carried out according to the number of armor of both sides, and then the OODA loop system adjusts the judgment and decision time coefficients for the next confrontation cycle according to the results of the first cycle. Compared with traditional simulation methods that consider objective factors such as loss rate and support rate, the OODA-loop-based hierarchical game analysis can analyze the confrontation situation more comprehensively.

\end{abstract}


\begin{multicols}{2}
\section{Introduction}
Modern warfare situations are fluid and require rapid ad hoc countermeasure decisions. Military technology is also advancing with the development of artificial intelligence, and the future warfare can be said to be dominated by the participation of few combatants and the unmanned nature of actual combat. Here to use NetLogo software simulation to build a dynamic projection of both sides tanks in complex terrain confrontation thought\cite{b1}, both sides confrontation terrain environment as shown in Figure 1.
\begin{figure*}[t]
\centerline{\includegraphics[width=6.7in]{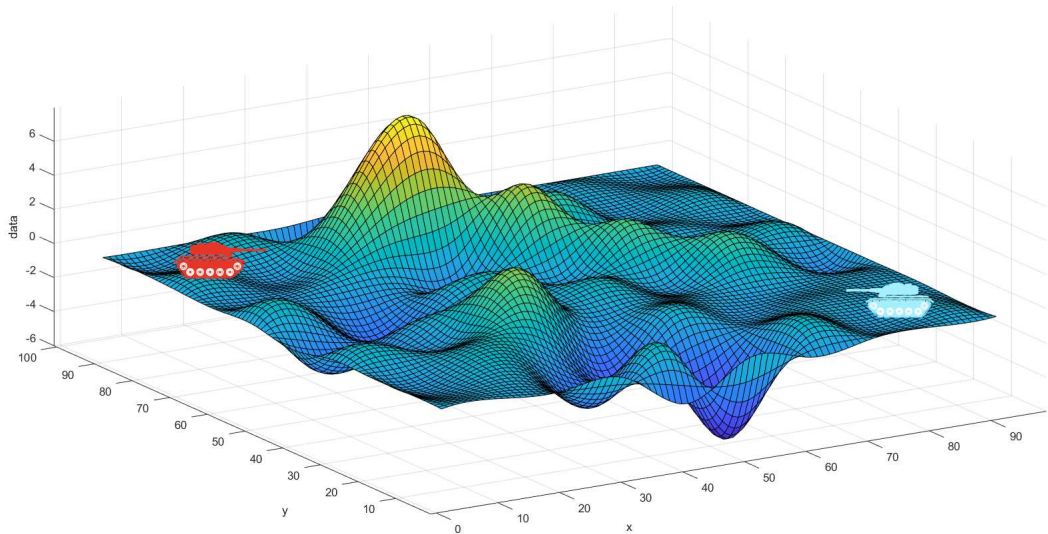}}
\caption{Diagram of the Game Environment of both Sides.}
\label{fig1}
\end{figure*}

The dynamic projection of the two sides' tank confrontation in complex terrain is conceived as a situation where the Red side is preparing to achieve a strike against the Blue side's chain of command in the complex terrain of the mountain peaks and valleys, and the Blue side intends to send a significant portion of its armor numbers to intervene in order to achieve the objective of stopping the strike and maintain the intention of containing the Red side's course. After the Red side learns of the Blue side's strategic intent, it sends its absolute superior armor numbers to execute counter-intervention operations at the first opportunity. The OODA cycle connects the two sides in the confrontation process, forming a complete game process, and finally using hierarchical analysis to provide a reference for the weight value of the mixed game strategy. -Decision-action" cycle. The "Orientation" and "Decision" steps are the most critical in the OODA cycle, and if the enemy's threat is misjudged or the surrounding environment is misunderstood, then the judgment and decision will be wrong\cite{b2,b3,b4,b5}. The goal is to interfere with the enemy's OODA cycle by taking prompt action after the first OODA cycle is completed by the own side. In constructing a complex combat system adversarial analysis, NetLogo modeling software will be used to model the system dynamics of the game process on both sides\cite{b6,b7,b8,b9,b10,b11}.

\section{Complex system confrontation model construction theory}
In the 1970s, the U.S. Air Force Colonel Boyd proposed the "$OODA$ Loop" theory. $OODA$ is the abbreviation of the four English words Observation, Orientation, Decision, and Action. Its meaning is to observe, Orientation, make decisions, and Action.\cite{b11,b12}The OODA Loop theory (Observation, Orientation, Decesion, Action) proposed by John Boyd provides a way to describe conflicts in the cycle of detection, judgment, decision-making, and action. Decision-making processing is widely used in the study of military issues. The OODA loop theory points out that the characteristic of command and control operations is to fight in accordance with the observation-judgment-decision-action loop, which can be divided into the following four stages: The first stage is observation (Observation), mainly through various sensing equipment and The use of the network for intelligence collection, for example, early warning detection information, battlefield environment information, enemy coordinate information, etc. The second stage is Orientation. The battlefield situation is changing rapidly and there is great uncertainty. Judgment is to convert data into useful information. Effective and rapid judgment of the results can assist the commander in making correct decisions and even changes. The battlefield situation. The third stage is Decision, where the commander clearly formulates a mission plan and issues a battle plan. The rapid formation of decision-making is of decisive significance to the battlefield situation. The fourth stage is action. Actions on the battlefield are usually described as taking corresponding operational deployments and performing actions based on the operational plan issued by the superior.

The OODA Loop has periodicity and nesting. Periodicity means that when a loop ends, another new loop begins. The size of the period is related to the scale of combat forces, spatial scope, and combat style; nesting means that the loop and The loops are related in the form of nesting. For example, in the tactical combat system, the smallest OODA Loop is the firepower closed-loop control loop of the close-range weapon system. There are also OODA Loops at the unit level. These command and control loops are nested within each other, and the inner loop The cycle is short and the outer loop cycle is long.

In combat, the opposing parties will continuously observe the surrounding environment, obtain relevant information, judge threats, make immediate adjustments, make decisions, and take corresponding actions. This is specifically reflected in the competition of OODA Loop response capabilities. If they can try to shorten their own side The enemy's OODA Loop and increase the enemy's OODA Loop as much as possible. New actions are initiated before the enemy responds to its previous action, causing the enemy to lose the ability to react, so that the initiative can be mastered and a huge advantage can be obtained..The schematic diagram of the $OODA$ Loop is shown in Figure 2.
\begin{figure*}[t]
\centerline{\includegraphics[width=5.0in]{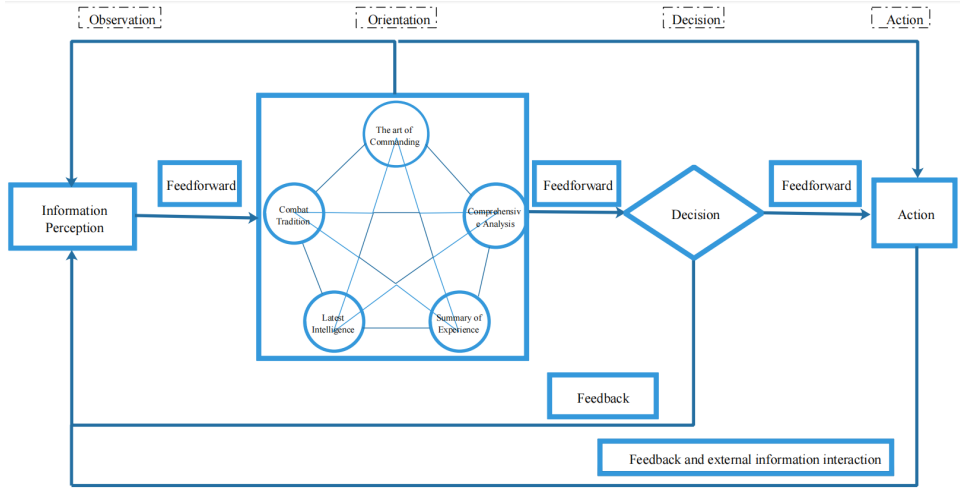}}
\caption{Diagram of the Game Environment of both Sides.}
\label{fig2}
\end{figure*}

The first is observation. Before the implementation of command operations, a clear observation plan is required. The two sides started by observing the new situation from the beginning of the war and the reaction of the other side. In the experiment, the red and blue sides were simulated and designed for the $OODA$ Loop, and the deployment was judged by observing the strength of the opposing sides before the war. The second is to judge and observe the situation and make reasonable decisions. In order to reflect the process of dynamically deploying the amount of armor, the slider value adjustment in the judgment and decision-making link in $NetLogo$ is used. At the end, the actual output of the red and blue sides is used as the execution link, and the amount of armor that needs to be dispatched for the next $OODA$ Loop is made according to the previous three links.\cite{b13,b14,b15}

The $OODA$ Loop is a closed ring for decision-making and control actions, and it has the characteristics of a periodic cycle. The law of future warfare is that combat units that make quick judgments and decisions will take the initiative. From the $OODA$ Loop theory, it can be concluded that the timeliness of military operations research. Each time period of the system combat command cycle depends on different factors. $\lambda_{d}$ and $\lambda_{f}$ are defined separately as the mobile deployment rhythm and engagement rhythm of the combat system in the physical domain.$\lambda_{C 2}$ is the decision-making rhythm of the combat system in the cognitive domain,$\lambda_{T}$is the rhythm of interaction between the information domain, the physical domain, and the cognitive domain,$\Delta t_{1}, \Delta t_{2}, \Delta t_{3}, \Delta t_{4}$is defined as the combat time of the observation, judgment, decision-making and action links of the OODA Loop,then complete the OODA Loop and the time required for each link, as shown in Equation 1 and Equation 2.
\begin{equation}\label{eq1}
\left\{\begin{array}{l}
\Delta t_{1} \geq \frac{1}{\lambda_{T}} \\
\Delta t_{2} \geq \frac{1}{\lambda_{T}}+\frac{1}{\lambda_{C 2}} \\
\Delta t_{3} \geq \frac{1}{\lambda_{T}}+\frac{1}{\lambda_{d}} \\
\Delta t_{4} \geq \frac{1}{\lambda_{T}}+\frac{1}{\lambda_{f}}
\end{array}\right.
\end{equation}

\begin{equation}\label{eq2}
\begin{aligned}
T_{O O D A} &=\Delta t_{1}+\Delta t_{2}+\Delta t_{3}+\Delta t_{4} \\
& \geq \frac{4}{\lambda_{T}}+\frac{1}{\lambda_{C 2}}+\frac{1}{\lambda_{d}}+\frac{1}{\lambda_{f}}
\end{aligned}
\end{equation}

According to Shannon's information entropy theory, the probability of occurrence of each part of the observation, judgment, decision-making and action links in the OODA Loop is defined as $P(O 1), P(O 2), P(D), P(A)$.Then the information entropy of each part of the observation, judgment, decision and action links in the OODA Loop is shown in Equation 3.

\begin{scriptsize}
\begin{equation}\label{eq3}
\left\{\begin{array}{l}
H\left(O 1_{i}\right)=-\sum_{i=1}^{n} P\left(O 1_{i}\right) \log \left(P\left(O 1_{i}\right)\right) \\
H\left(O 2_{i}\right)=-\sum_{i=1}^{n} P\left(O 2_{i}\right) \log \left(P\left(O 2_{i}\right)\right) \\
H\left(D_{i}\right)=-\sum_{i=1}^{n} P\left(D_{i}\right) \log \left(P\left(D_{i}\right)\right) \\
H\left(A_{i}\right)=-\sum_{i=1}^{n} P\left(A_{i}\right) \log \left(P\left(A_{i}\right)\right)
\end{array}\right.
\end{equation}
\end{scriptsize}
In the formula, n indicates that the OODA Loop can be multiple cycles.

The information entropy of all OODA Loop of complex operations systems is shown in formula 4.
\begin{scriptsize}
\begin{equation}\label{eq4}
\begin{aligned}
H(O O D A) &=H\left(O 1_{i}\right)+H\left(O 2_{i}\right)+H(D)+H(A) \\
&=-\sum_{i=1}^{n} P\left(O O D A_{i}\right) \log \left(P\left(O O D A_{i}\right)\right)
\end{aligned}
\end{equation}
\end{scriptsize}

\emph{Statement: According to the information entropy of each link of the OODA Loop, the information entropy of the OODA Loop of the complex combat system can be described. An OODA Loop is composed of four links to solve the maximum information entropy of all OODA Loop in a complex combat system.The information entropy of all OODA Loop of the system is $H(O O D A)=-\sum_{i=1}^{n} P\left(O O D A_{i}\right) \log \left(P\left(O O D A_{i}\right)\right)$,Then prove $H(O O D A) \leq \log _{2} n$.}

\begin{proof}
The information entropy of the complex system is abbreviated as Equation 5, and the constraint condition is Equation 6. The problem of solving the maximum value of the original Equation 4 is equivalent to the minimum value of Equation 5..
\end{proof}

\begin{equation}\label{eq5}
H(D)=\sum_{i=1}^{n} X_{i} \log _{2} X_{i} \\
\end{equation}

\begin{equation}\label{eq6}
\sum_{i=1}^{n} X_{i}=1
\end{equation}

Then construct the Lagrangian function as formula 7
\begin{scriptsize}
\begin{equation}\label{eq7}
L\left(X_{1}, \cdots \cdots, X_{n}\right)=\sum_{i=1}^{n} X_{i} \log _{2} X_{i}+\lambda\left(\sum_{i=1}^{n} X_{i}-1\right)
\end{equation}
\end{scriptsize}

\begin{tiny}
\begin{equation}
\begin{aligned}
\frac{\partial L\left(X_{1}, \cdots \cdots, X_{n}\right)}{\partial X_{1}} &=\frac{\partial}{\partial X_{1}}\left[\sum_{i=1}^{n} X_{i} \log _{2} X_{i}+\lambda\left(\sum_{i=1}^{n} X_{i}-1\right)\right]=0 \\
&=\log _{2} X_{1}+\frac{1}{\ln 2}+\lambda=0 \\
& \Rightarrow \lambda=-\log _{2} X_{1}-\frac{1}{\ln 2} \\
\frac{\partial L\left(X_{1}, \cdots \cdots, X_{n}\right)}{\partial X_{2}} &=\frac{\partial}{\partial X_{2}}\left[\sum_{i=1}^{n} X_{i} \log _{2} X_{i}+\lambda\left(\sum_{i=1}^{n} X_{i}-1\right)\right]=0 \\
&=\log _{2} X_{2}+\frac{1}{\ln 2}+\lambda=0 \\
& \Rightarrow \lambda=-\log _{2} X_{2}-\frac{1}{\ln 2}
\end{aligned}
\end{equation}

......

\begin{equation}
\begin{aligned}
\frac{\partial L\left(X_{1}, \cdots \cdots, X_{n}\right)}{\partial X_{n}} &=\frac{\partial}{\partial X_{n}}\left[\sum_{i=1}^{n} X_{i} \log _{2} X_{i}+\lambda\left(\sum_{i=1}^{n} X_{i}-1\right)\right]=0 \\
&=\log _{2} X_{n}+\frac{1}{\ln 2}+\lambda=0 \\
& \Rightarrow \lambda=-\log _{2} X_{n}-\frac{1}{\ln 2}
\end{aligned}
\end{equation}

\begin{equation}
\begin{aligned}
\frac{\partial L\left(X_{1}, \cdots \cdots, X_{n}\right)}{\partial \lambda}=& \frac{\partial}{\partial \lambda}\left[\sum_{i=1}^{n} X_{i} \log _{2} X_{i}+\lambda\left(\sum_{i=1}^{n} X_{i}-1\right)\right]=0 \\
& \Rightarrow \sum_{i=1}^{n} X_{i}=1
\end{aligned}
\end{equation}
\end{tiny}
The above equation can be solved.
\begin{equation}
X_{1}=X_{2}=\ldots=X_{n}=\frac{1}{n}
\end{equation}

So $H(O O D A)=-\sum_{i=1}^{n} \frac{1}{n} \log _{2} \frac{1}{n}=\log _{2} n$.

The information entropy value of all OODA Loop of a complex combat system is a system that increases as the number of system cycles increases, indicating that the system is a highly complex entropy increasing system and the entropy value is in a state of divergence. For the highly complex entropy increase system, Swarm, $NetLogo$ and other software are needed to assist the analysis. $NetLogo$ has the characteristics of simple model description language and rapid modeling. In the third section, $NetLogo$ modeling specifically analyzes the process of confronting both parties.

\section{Netlogo modeling based on OODA Loop}

$Netlogo$ is a programmable modeling environment used to simulate natural and social phenomena.\cite{b16,b17,b18}When the internal operating rules of a complex system are not obvious and it is difficult to grasp the rules, $NetLogo$ software can be used to simulate the system.\cite{b19}The red and blue confrontation experiment simulation is based on the hardware platform of $Inter(R) Core(TM) i7-10750H CPU @2.60GHZ$ and the software platform of $NetLogo 6.1.1$. Taking the construction of the red OODA Loop as an example, the modeling is carried out from four aspects: observation, judgment, decision-making and action.The research of complex nonlinear combat with multi-agent systems in modern combat is an important direction of war simulation. NetLogo is a powerful multi-agent simulation tool, which greatly improves the simple analytical model and analytical combat methods. Due to the simplification of the research, this article limits the combat process to the micro process of OODA Loop combat. In the future, the simulation can be extended and expanded to achieve OODA Loop stacking and cumulative confrontation. Use NetLogo to improve the level of human-computer interaction, establish models and calculation formulas with a higher degree of reduction, and analyze the process and results of simulation operations to draw more credible evaluation conclusions. By analyzing the problems, finding the gaps, laying the foundation for the formulation of combat plans, so that simulation can truly serve the combat decision-making. The functions of each module in the OODA Loop are defined as follows:
Based on practical foundation and facts, the difference in the amount of armor between the two opposing parties is relatively large, so the party with the less armor will take a longer time to think and need more time to deploy armor to achieve the goal of winning with less. Select the value of the $NetLogo$ slider for the judgment and decision-making links of the red and blue parties. For example, 500 armors of the red side and 400 armors of the blue side appear at one of the nodes of the OODA Loop, and Then the red and blue sides in the NetLogo simulation software slider affiliation value is adjusted to 1 and 0.5 respectively. as shown in Figure 3, Figure 4.

\begin{figurehere}
\begin{center}
\centerline{\includegraphics[width=2in]{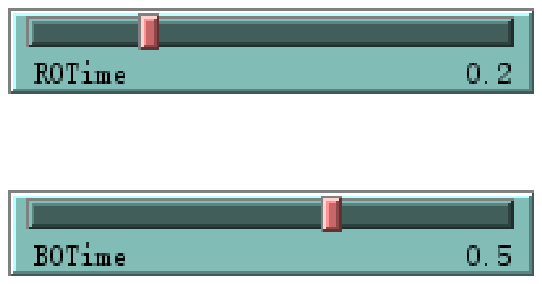}}
\caption{Time adjustment module of the judgment link.}
\label{fig3}
\end{center}
\end{figurehere}

\begin{figurehere}
\begin{center}
\centerline{\includegraphics[width=2in]{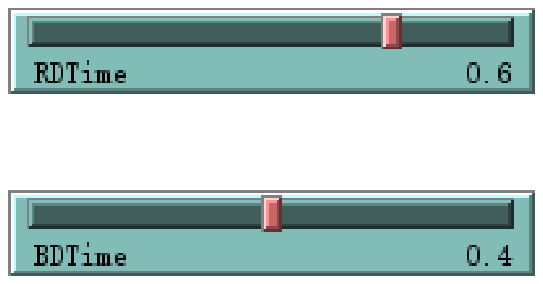}}
\caption{Time adjustment module for the decision-making link.}
\label{fig4}
\end{center}
\end{figurehere}

In order to quickly select $ROTime$, $BOTime$, $RDTime$, and $BDTime$ values, some judgment and decision time values are now listed, as shown in $Table 1$.
\end{multicols}

\begin{tablehere}
\tbl{OODA combat Loop part Orientation and Decision-making time value reference table. \label{tab1}}
{\begin{tabular*}{16.4cm}{ccccc}
	\toprule
\begin{tabular}[c]{@{}c@{}}Red and Blue \\ Armmored Quantity(X,Y)\end{tabular} &
  \begin{tabular}[c]{@{}c@{}}Red Determine the \\ time value\end{tabular} &
  \begin{tabular}[c]{@{}c@{}}Blue Determine the \\ time value\end{tabular} &
  \begin{tabular}[c]{@{}c@{}}Red Decision \\ time value\end{tabular} &
  \begin{tabular}[c]{@{}c@{}}Blue Decision \\ time value\end{tabular} \\
  \colrule
  (500,200) & 0.3 & 1.0 & 0.3 & 1.0 \\
  (500,300) & 0.4 & 1.0 & 0.4 & 1.0 \\
  (500,400) & 0.4 & 0.7 & 0.4 & 0.7 \\
  (600,200) & 0.2 & 1.0 & 0.2 & 1.0 \\
  (600,300) & 0.3 & 1.0 & 0.3 & 1.0 \\
  (600,400) & 0.2 & 0.5 & 0.2 & 0.5 \\
  (700,200) & 0.1 & 1.0 & 0.1 & 1.0 \\
  (700,300) & 0.1 & 0.9 & 0.1 & 0.9 \\
  (700,400) & 0.2 & 0.7 & 0.2 & 0.7 \\
	\botrule
\end{tabular*}}
\end{tablehere}
\begin{multicols}{2}
\subsection{Investigation link}
The reconnaissance behavior model mainly describes the combat behavior of various detection, reconnaissance equipment and various sensing devices, which is the main means to obtain intelligence information on the battlefield. In the OODA Loop, the information acquisition of the investigation link is related to the pre-war deployment, command and decision-making, etc. The Red investigation consists of three modules, the speed at which the Red obtains intelligence ($RIS$), the speed preventing the Red from acquiring intelligence ($BZuA$), and the speed the Red actually obtains intelligence ($R-Actual-S$). The system dynamics module composed of the three modules is as shown in Figure 5.
\begin{figurehere}
\begin{center}
\centerline{\includegraphics[width=2in]{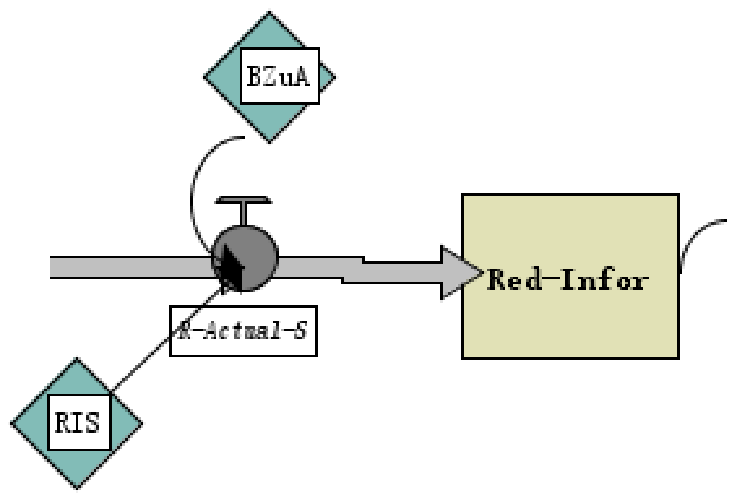}}
\caption{Dynical model of Red Square investigation link.}
\label{fig5}
\end{center}
\end{figurehere}
\subsection{Judgment link}
This part of the OODA Loop is to analyze the collected intelligence, determine the authenticity of the intelligence, and study the current situation. It is mainly completed by the red judgment ($RO$) and the red judgment time ($ROTime$) in the combat elements of the command system. It is an accusation The processing center of the information flow in the system. The red side judgment link is composed of two modules, namely, red side judgment ($RO$) and red side judgment flow input ($RIn$). The system dynamics module composed of two modules is shown in Figure 6.
\begin{figurehere}
\begin{center}
\centerline{\includegraphics[width=2in]{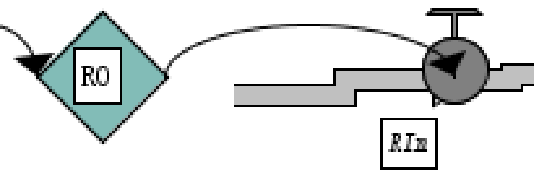}}
\caption{Dynical model of red square judgment link.}
\label{fig6}
\end{center}
\end{figurehere}
\subsection{Decision link}
The decision-making link in the OODA Loop is based on the situation judgment, assigning the action and task plan, mainly by the red decision ($RD$) module in the combat elements of the accusation system, generating a new command and control information flow. The decision link consists of red temporary storage module ($RlinshiCun$) and red decision module ($RD$). The system dynamics model of the decision-making link is shown in Figure 7.
\begin{figurehere}
\begin{center}
\centerline{\includegraphics[width=1in]{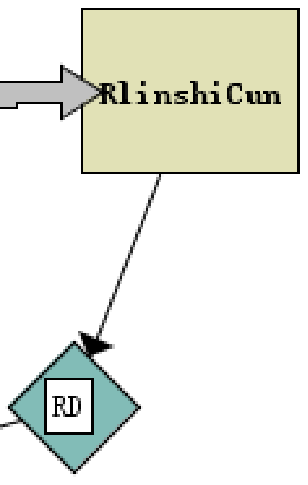}}
\caption{Red square decision link dynamics model.}
\label{fig7}
\end{center}
\end{figurehere}
\subsection{Action link}
The action link in the OODA Loop is the corresponding action behavior based on the decision made and the designated action plan, mainly through the red square armor module ($Reds$), the red side loss factor module ($RSunhaoxishu$), the red side actual output module ($Redsshijishuchu$), and the blue side Decision Module ($BD$),It is the specific implementation of the charge system information flow,as shown in Figure 8.
\begin{figurehere}
\begin{center}
\centerline{\includegraphics[width=1.5in]{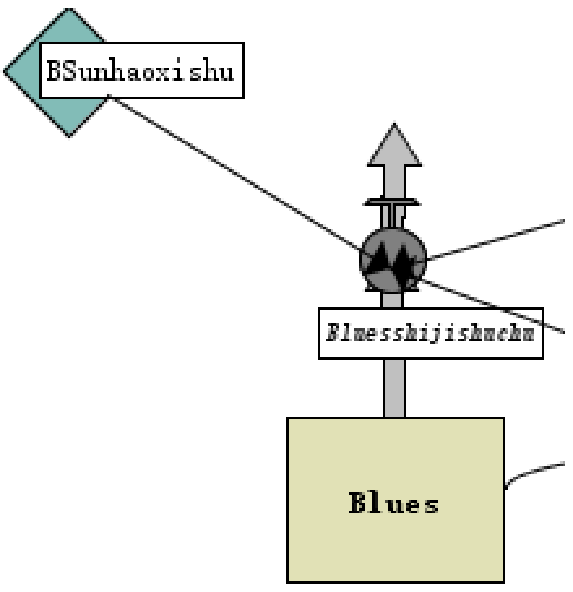}}
\caption{Red square action link dynamics model.}
\label{fig8}
\end{center}
\end{figurehere}
The dynamic model of the Observation, Orientation, Decision, Action link of the blue OODA Loop is shown in Figure 9 below.
\begin{figurehere}
\begin{center}
\begin{tabular}{c}
\centerline{\includegraphics[width=2in]{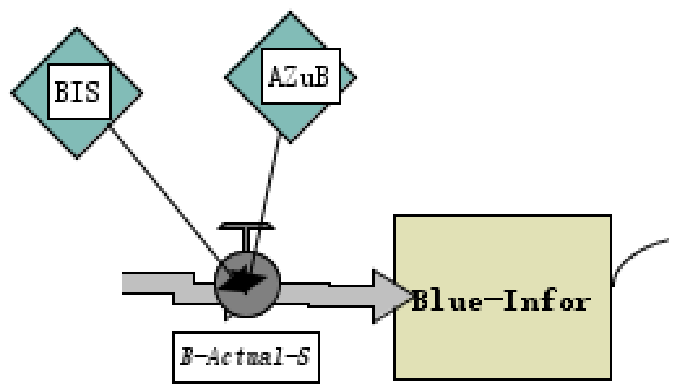}}\\
{\footnotesize\sf {(a)Dynical model of blue fang investigation link.}}\\
\centerline{\includegraphics[width=2in]{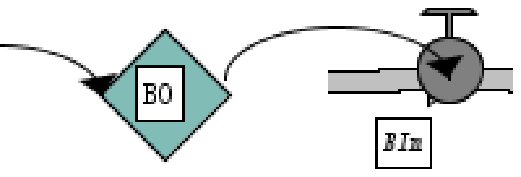}}\\
{\footnotesize\sf {(b)Blue side judgment link dynamics model.}}\\
\centerline{\includegraphics[width=1in]{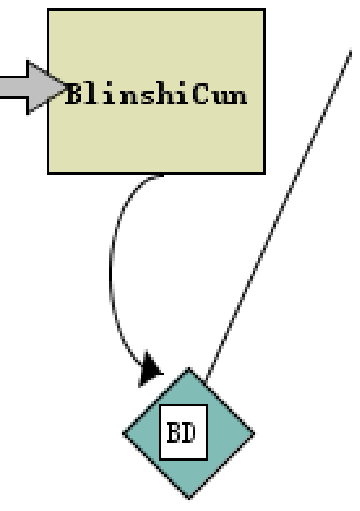}}\\
{\footnotesize\sf {(c)Blue square decision-making link dynamics model.}} \\
\centerline{\includegraphics[width=2in]{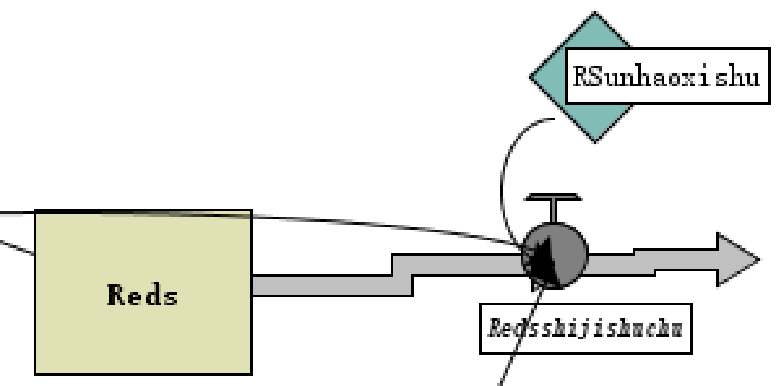}}\\
{\footnotesize\sf {(d)Blue square action link dynamics model.}} \\
\end{tabular}
\caption{The logic pattern of Blue with OODA loop.}
\end{center}
\end{figurehere}
\section{Traditional methods are compared with OODA Loop}\label{sec4}
The simple system uses two influencing factors as the OODA Loop of NetLogo simulation. Compared analysis shows the difference between the two methods and whether OODA Loop can improve operational thinking.
\subsection{Simple system}
Assuming the amount of red and blue armor, the two sides fight according to certain rules of engagement. Both parties then adjust their armor numbers and operational capabilities to simulate their operations. The kinetic model of a simple system is shown in $Fig.10$ and $Fig.11$.The combat system is a complex system composed of personnel, equipment, and organization. The combat process is affected by many factors such as combat missions, technical support, logistical support, suddenness, morale, and command art. It is very difficult and tedious to comprehensively analyze and model the factors in the combat process. Therefore, the combat process is specifically abstracted as the link of dynamic combat interaction, and only the system process of red and blue tank confrontation is analyzed, which can also play a good simulation role. By adjusting the number and combat capabilities of the tanks on both sides, let them conduct simulation operations.

Simulate the number of tanks on both sides, represented by R-Tank and G-Tank. As the confrontation between the two warring parties continues, the number of tanks will show a certain consumption trend. If the number of subsequent tanks is supplemented, it will lead to a certain decrease or increase in the scale of the tanks of both sides. The confrontation between the two sides is a fierce process. The number of tanks is high. In order to ensure the continuity of combat effectiveness, continuous replenishment of tank armor is required. R-Input and G-Input are used to indicate the number of replenishment of the red and blue sides. At the same time, use R- lost and G-lost indicate the amount of equipment lost by both red and blue. There are four variables in this model, namely the loss rate of the red and blue tanks and the replenishment rate of the red and blue tanks. R-lost-rate, G-lost-rate, R-input-rate and G-input-rate are used respectively. Express. In the process of this model test, the Red Army tanks are replenishing (R-input) and losing (R-lost). The confrontation is carried out step by step. Due to the short process of modern warfare and the fierce confrontation, the speed of equipment replenishment is often lower than that of The speed at which weapons and equipment are consumed.
\begin{figurehere}
\begin{center}
\centerline{\includegraphics[width=3in]{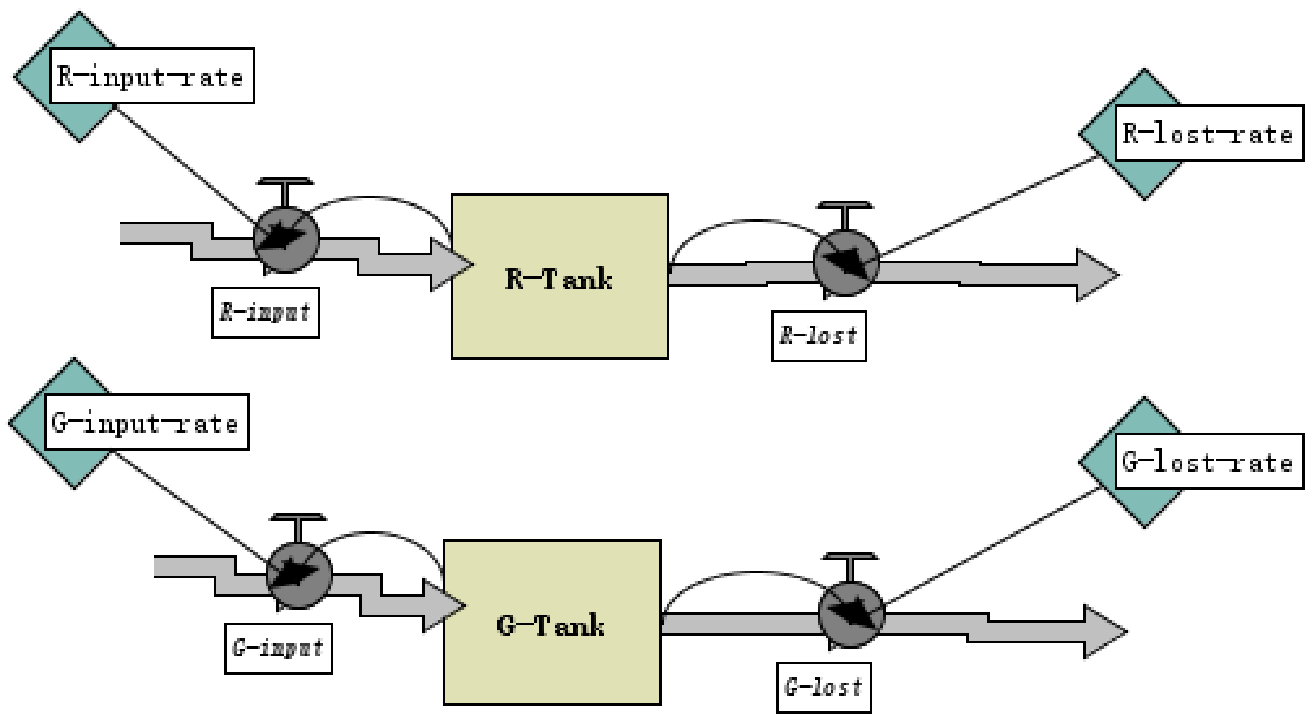}}
\caption{Schematic of simple system dynamics model.}
\label{fig10}
\end{center}
\end{figurehere}
\begin{figurehere}
\begin{center}
\centerline{\includegraphics[width=2.5in]{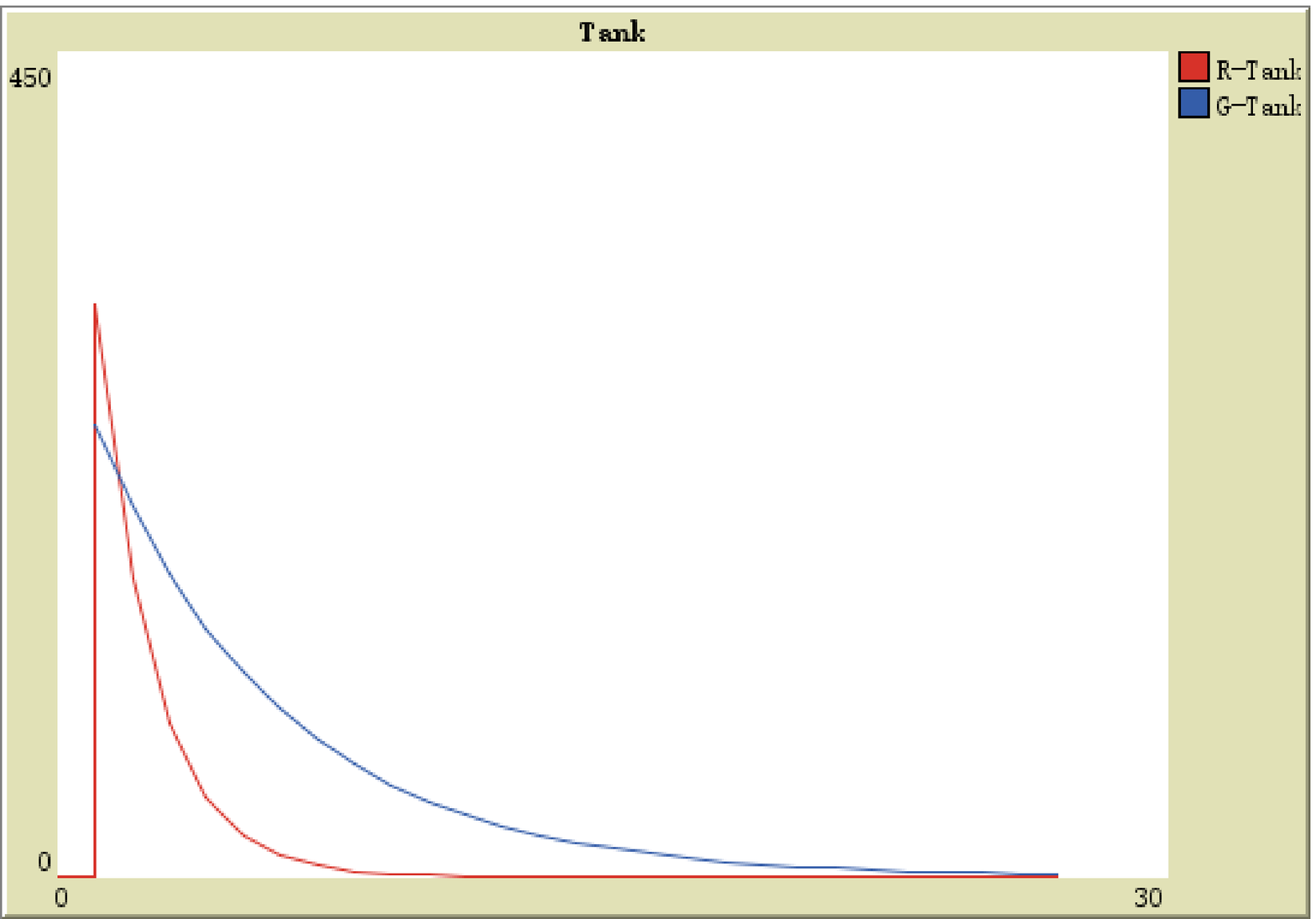}}
\caption{Schematic diagram of the red and blue confrontation results of the simple system.}
\label{fig11}
\end{center}
\end{figurehere}
\subsection{OODA Loop system}
Through simulation of red and blue OODA Loop by NetLogo in Fig.5-Fig.9, the dynamic model of complex combat system is shown in Figure 12, and the schematic diagram of red and blue confrontation results of OODA Loop system is shown in Figure 13.
\begin{figure*}[tbh]
	\centering
	\includegraphics[scale=0.6]{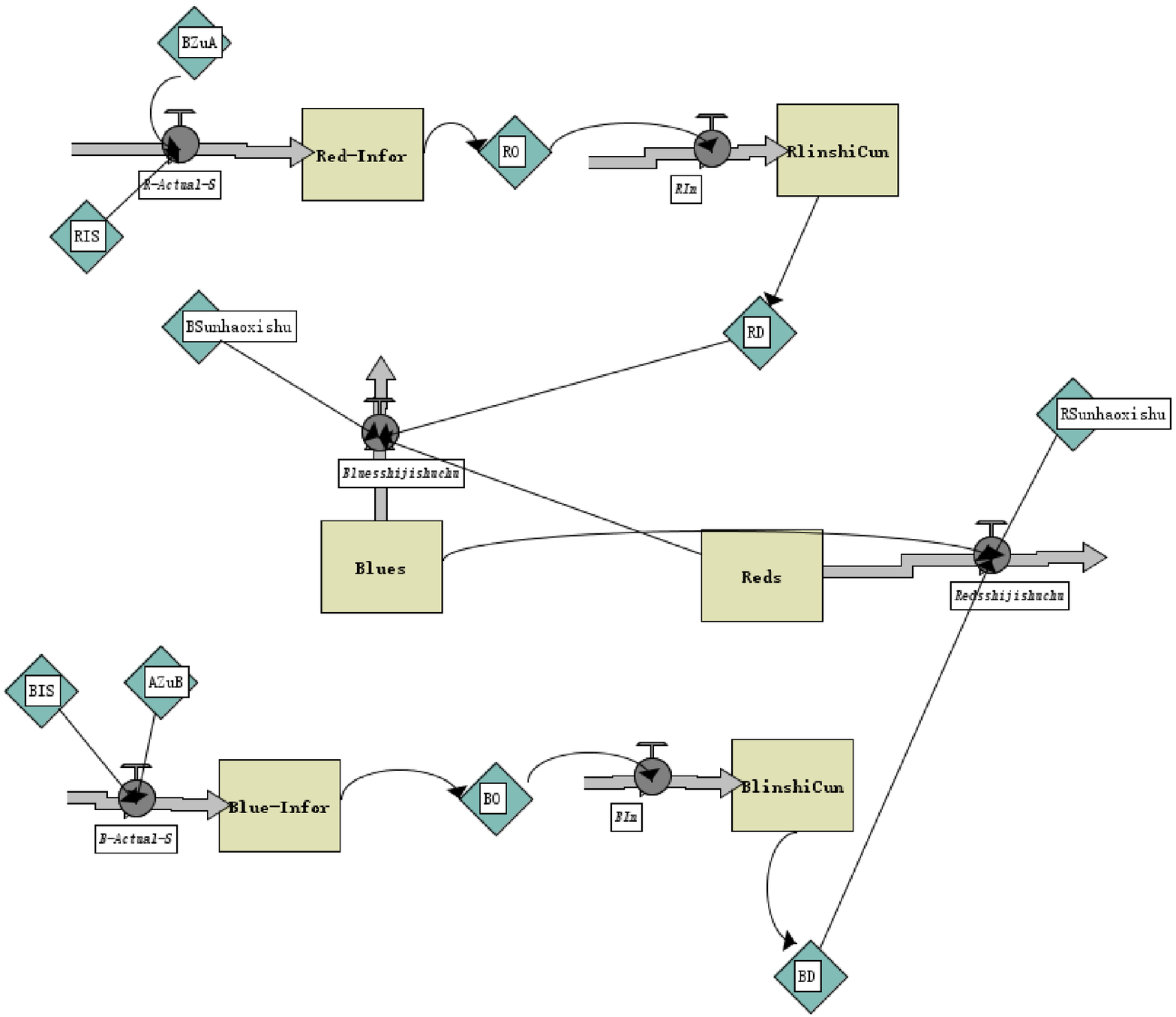}
	\caption{Dynamical models of complex combat systems based on OODA Loop.} 
\end{figure*}
\begin{figurehere}
\begin{center}
\centerline{\includegraphics[width=2.5in]{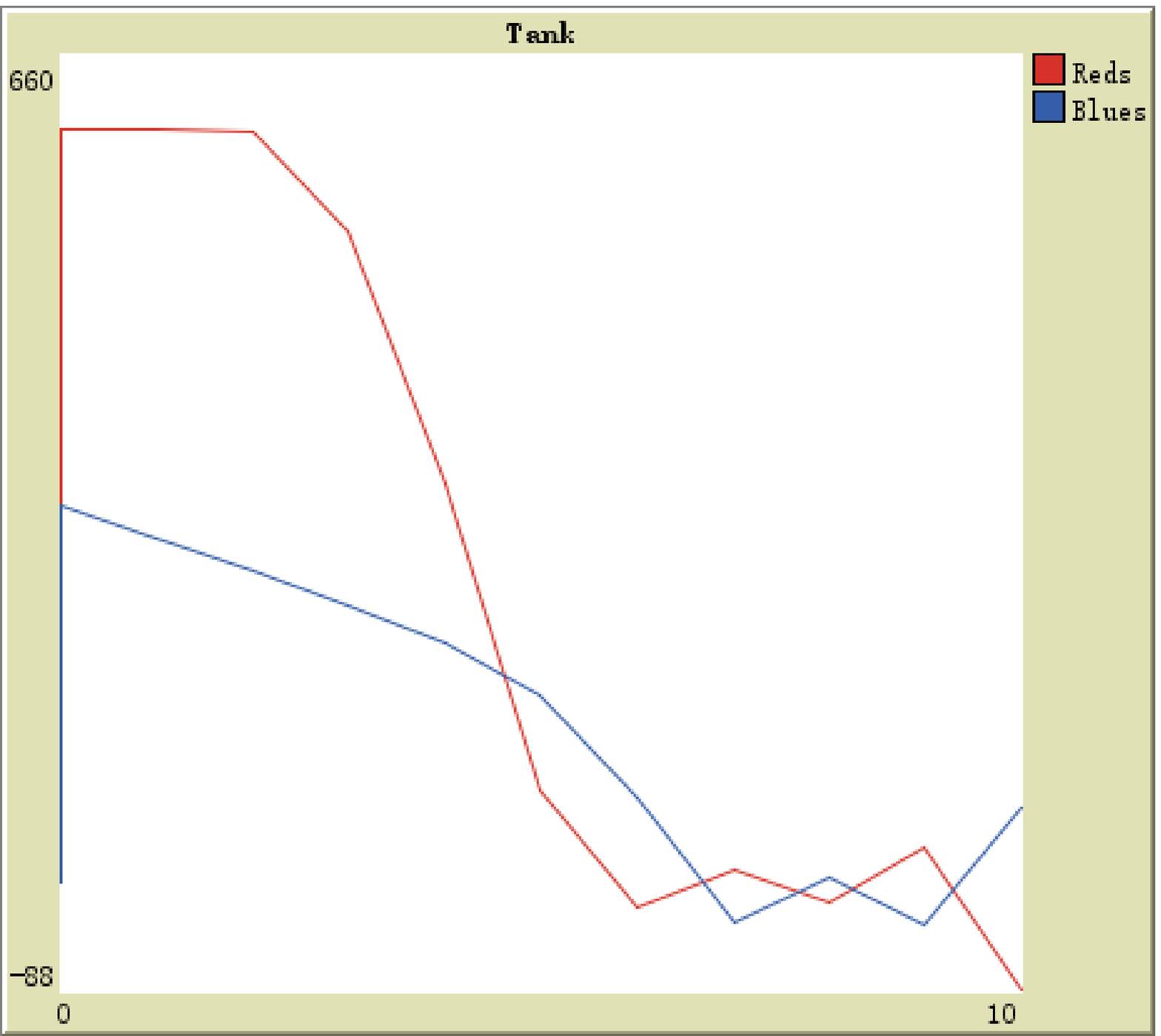}}
\caption{Schematic diagram of the red and blue confrontation results of the OODA Loop system.}
\label{fig13}
\end{center}
\end{figurehere}
\subsection{OODA Loop system feasibility verification}
The OODA cycle complex OS scheme is selected according to the Operations Research Hierarchy Analysis Method (The Analytic Hierarchy Process, or AHP). Hierarchical analysis is a systematic and quantitative approach proposed by operations strategist Professor T. L. Saaty in the early 1970s.

The analytic hierarchy process refers to a complex multi-objective decision-making problem as a system, the goal is decomposed into multiple goals or criteria, and then decomposed into multiple indicators (or criteria, constraints) several levels, through the qualitative index fuzzy quantitative method to calculate the level Single ranking (weight) and total ranking are used as a systematic method for target (multi-index) and multi-scheme optimization decision-making.The analytic hierarchy process is to decompose the decision-making problem into different hierarchical structures in the order of the overall goal, the sub-goals of each level, the evaluation criteria, and the specific investment plan, and then use the method of solving the eigenvectors of the judgment matrix to obtain each level The priority of an element to an element of the previous level, and finally the weighted sum method is recursively merged and the final weight of each alternative plan to the overall goal, and the one with the largest final weight is the optimal plan.The analytic hierarchy process is more suitable for decision-making problems with hierarchical and interlaced evaluation indicators, and the target value is difficult to describe quantitatively.\cite{b20}

First OODA Loop complex CS scheme hierarchical hierarchy, as shown in Figure 14. The hierarchical hierarchy includes three levels, target, criterion and scheme. The target layer selects options for complex combat systems; the criterion layer includes four modules: reconnaissance, judgment, decision, and execution; the program layer includes three schemes: support, continuation and retreat.
\begin{figurehere}
\begin{center}
\centerline{\includegraphics[width=2.5in]{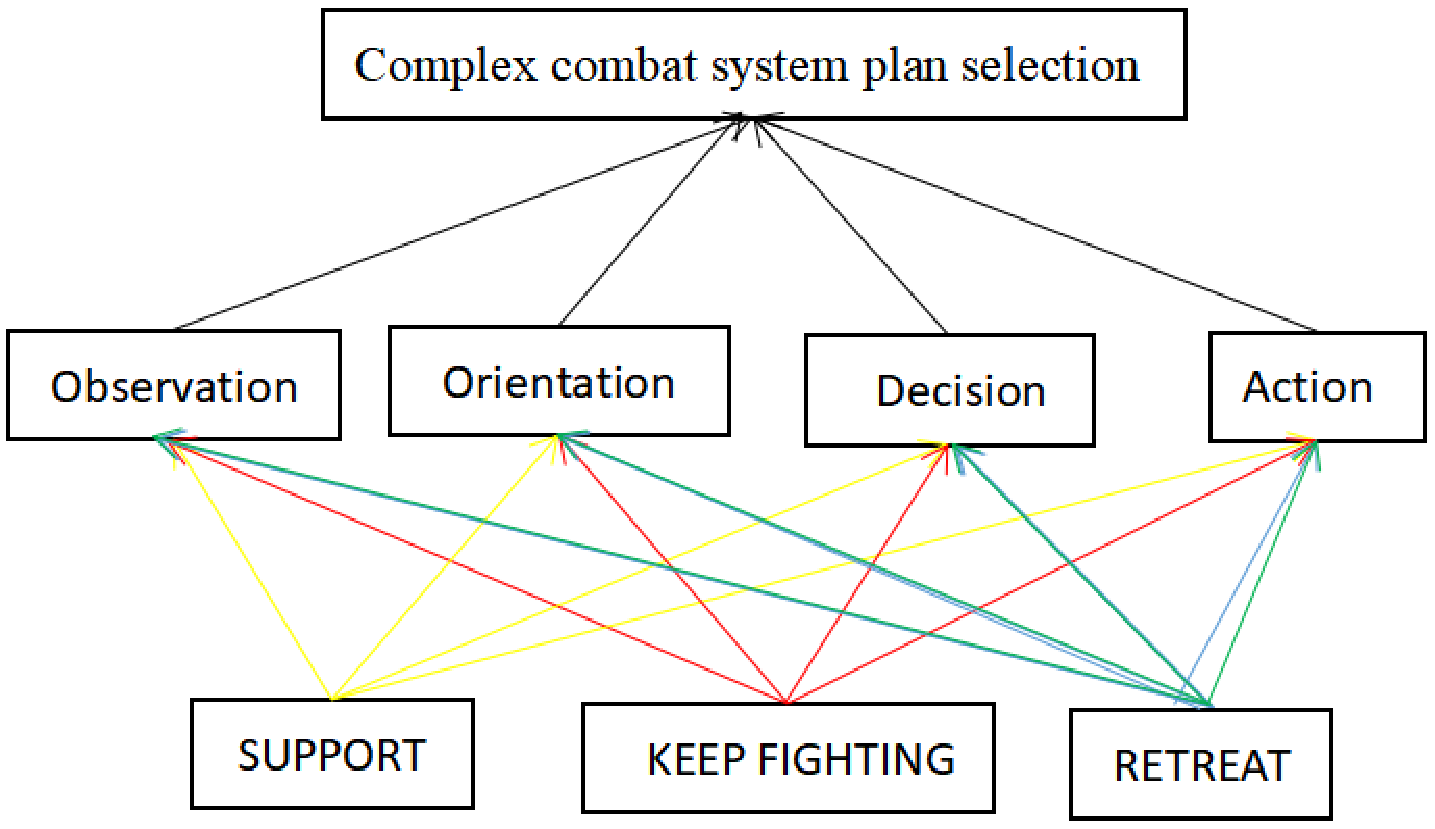}}
\caption{Hierarchical structure of OODA Loop complex combat system plan.}
\label{fig14}
\end{center}
\end{figurehere}
Secondly, according to the scale value table of the analytic hierarchy process, the consistency matrix of the complex combat system plan to the four criterion modules (as shown in $Table 2$ and $Table 3$) and the consistency matrix of the three combat plans to each part of the OODA Loop are constructed. Due to space limitations, here only the consistency matrix of the four criterion modules of the complex combat system plan and the selection weight matrix of the complex combat system plan (as shown in Table 4) are constructed here, and the consistency of the three combat plans for each part of the OODA Loop The matrix will be explained in the appendix.
botrule
\end{multicols}

\begin{tablehere}
\tbl{The scale value matrix of the four criterion modules of the complex combat system plan.\label{tab2}}
{\begin{tabular}{ccccc}
\toprule
\begin{tabular}[c]{@{}c@{}}Complex combat system \\ plan selection(A)\end{tabular} & Observation & Orientation & Decision & Action \\ \colrule
Observation          & 1.0000 & 3.0000 & 4.0000 & 7.0000  \\
Orientation          & 0.3333 & 1.0000 & 2.0000 & 3.0000  \\
Decision             & 0.2500 & 0.5000 & 1.0000 & 2.0000  \\
Action               & 0.1429 & 0.3333 & 0.5000 & 1.0000  \\
Sum of scaled values & 1.7262 & 4.8333 & 7.5000 & 13.0000 \\ \botrule
\end{tabular}}
\end{tablehere}

\begin{tablehere}
\tbl{Consistency matrix of complex combat system plan to four criterion modules.\label{tab3}}
{\begin{tabular}{ccccccc}
\toprule
\begin{tabular}[c]{@{}c@{}}Complex combat system \\ plan selection\end{tabular} & Observation & Orientation & Decision & Action & Weights(w) & Aw \\ \colrule
Observation & 0.5793 & 0.6207 & 0.5333 & 0.5385 & 0.5679 & 2.2933 \\
Orientation & 0.1931 & 0.2069 & 0.2667 & 0.2308 & 0.2244 & 0.9029 \\
Decision    & 0.1448 & 0.1034 & 0.1333 & 0.1538 & 0.1339 & 0.5357 \\
Action      & 0.0828 & 0.0690 & 0.0667 & 0.0769 & 0.0738 & 0.2967 \\ \botrule
\end{tabular}}
\end{tablehere}
\begin{multicols}{2}
According to the above table data, the $Expert Choice 11.5$ software is used to calculate the value of its natural attributes to judge whether the matrix construction conforms to logical consistency. Among them, the characteristic maximum value $\lambda_{\max }$=4.0206, $CI=0.0069$, $RI=0.89$, $CR=0.0077$, since $CR<0.1$, the consistency matrix is logical.
\end{multicols}
\begin{tablehere}
\tbl{Selection Weight Matrix of Complex Combat System Scheme.\label{tab4}}
{\begin{tabular}{ccccc}
\toprule
\begin{tabular}[c]{@{}c@{}}Complex combat system \\ scheme selection\end{tabular} &
  \begin{tabular}[c]{@{}c@{}}Comprehensive \\ weight ratio\end{tabular} &
  \begin{tabular}[c]{@{}c@{}}Support \\ weight ratio\end{tabular} &
  \begin{tabular}[c]{@{}c@{}}Continue with the \\ operational weight ratio\end{tabular} &
  \begin{tabular}[c]{@{}c@{}}Retreback \\ weight ratio\end{tabular} \\ \colrule
Observation       & 0.5679 & 0.4577 & 0.1263 & 0.4160 \\
Orientation       & 0.2244 & 0.5571 & 0.3202 & 0.1226 \\
Decision          & 0.1339 & 0.5571 & 0.3202 & 0.1226 \\
Action            & 0.0738 & 0.7015 & 0.2267 & 0.0718 \\
\begin{tabular}[c]{@{}c@{}}Weighted \\ weight ratio\end{tabular} &        & 0.5113 & 0.2032 & 0.2855 \\ \botrule
\end{tabular}}
\end{tablehere}
\begin{multicols}{2}
According to the selection weight matrix of the complex combat system scheme, it can be concluded that in the simulation combat, it is not the mode described by the simple system. The two sides are simply fighting, and the amount of armor continues to decline; instead, through the description of the combat mode like the OODA Loop, the two sides Confrontation is a reciprocating process. At the same time, the dynamic process of the zigzag fluctuation of the armor quantity shown in the schematic diagram of the confrontation between the red and blue sides of the OODA Loop system is verified from the side.

It can be concluded from $Table 4$ that the weight ratios of the three combat plans are respectively, the support weight ratio is $0.5113$, and the continued combat weight and the retreat weight ratio are roughly the same. It shows that when using the OODA Loop to analyze the combat plan, the warring parties should support their own operations as much as possible, but the continued combat and retreat plan also accounted for nearly half of the proportion, and the choice to continue combat or retreat based on the confrontation situation.

\section{Conclusions}

This paper simulates a complex environment combat model based on OODA loop through NetLogo software and uses hierarchical analysis to embody the game strategy under three scenarios of attack, retreat, and continued reinforcement that the red side should take after the current OODA loop is executed in terms of the number of armor. It can be concluded from the experimental results that the simulation system of OODA loop is more exploratory and feasible compared to the traditional simple system approach simulation. The OODA cycle is a correct practice of modern military theory, and the full use of the OODA cycle can form a closed-loop control of the combat confrontation and make the combat analysis comprehensive and concrete.

\section*{Acknowledgments}
This research is partially supported by Nature Science Foundation of China with ID 62076028 and by National Key Research and Development Program of China with ID 2018AAA0101000.

\clearpage

\end{multicols}
\end{document}